\documentclass[letterpaper]{article} 
\usepackage[preprint]{aaai2027}  
\usepackage[hyphens]{url}  
\usepackage{graphicx} 
\usepackage{natbib}  
\usepackage{caption} 
\usepackage{algorithm}
\usepackage{algorithmic}
\usepackage{amsmath}
\usepackage{amsfonts}
\usepackage{booktabs}
\usepackage{multirow} 
\usepackage{tcolorbox}

\usepackage{newfloat}
\usepackage{listings}
\DeclareCaptionStyle{ruled}{labelfont=normalfont,labelsep=colon,strut=off} 
\floatstyle{ruled}
\newfloat{listing}{tb}{lst}{}
\floatname{listing}{Listing}

\usepackage{booktabs}
\usepackage{xcolor}
\definecolor{codegray}{rgb}{0.96,0.96,0.96}
\definecolor{commentgreen}{rgb}{0.1,0.6,0.1}
\definecolor{keywordblue}{rgb}{0.1,0.1,0.8}
\definecolor{errorred}{rgb}{0.8,0.1,0.1}

\newif\ifhighlight

\highlightfalse 

\newcommand{\xiang}[1]{%
  \ifhighlight
    \textcolor{orange}{#1}%
  \else
    #1%
  \fi
}
\title{Pin Once, Swap Light: Subspace-Aligned Centroid-Residual Training for Efficient Ultra-LoRA Serving}
\author {
    Xiang Li\textsuperscript{\rm 1},
    Pengcheng Wang\textsuperscript{\rm 1},
    Huazheng Wang\textsuperscript{\rm 2},
    Saurabh Bagchi \textsuperscript{\rm 1}
}
\affiliations {
    \textsuperscript{\rm 1}Purdue University\\
    \textsuperscript{\rm 2}Oregon State University
}

\newcommand{\name}{SALT}

\begin{document}

\maketitle

\begin{abstract}
Modern multi-tenant Low-Rank Adapters (LoRAs) serving systems concurrently host tens to hundreds of LoRA adapters. Though powerful, this introduces a critical system dilemma between serving efficiency and task performance: higher-rank adapters generally achieve better downstream task performance, but their GPU VRAM footprint and Host-to-Device PCIe swapping overhead severely constrain scalability. Conversely, ultra-low-rank adapters ($r \le 2$) minimize both VRAM footprint and PCIe transfer overhead, but suffer from downstream task performance degradation. To solve this problem, we propose Subspace-Aligned LoRA Training (\name), a serving efficiency-aware hierarchical fine-tuning framework. Our solution operates in three phases. First, a provider jointly trains high-capacity domain centroids on public data within the domain using a novel alignment regularizer that coheres in-domain task subspaces into a unified basis. Next, users fine-tune ultra-low-rank task residual adapters on private data atop those frozen centroids. Finally, during inference, the provider pins the centroid in GPU VRAM and dynamically swaps in each user's task residual on demand. Across LLMs of varying scales, \name\ recovers high-rank accuracy using $r \le 2$ residuals, achieving up to 18.5\% absolute accuracy gains over state-of-the-art compression baselines and reducing per-adapter memory by up to 16x. When integrated into vLLM, \name\ improves serving throughput by up to 51\% under PCIe bandwidth pressure and 28\% under GPU VRAM constraints for Llama-3.2-3B.

\end{abstract}


\section{Introduction}
\label{sec:intro}
Low-Rank Adaptation (LoRA)~\citep{hu2022lora,chen2024longlora,dettmers2023qlora} has transformed the fine-tuning landscape of Large Language Models (LLMs). By confining weight updates to low-rank matrices $A\in \mathbb{R}^{m \times r}$ and $B\in \mathbb{R}^{r \times  n}$, where $r\ll min(m,n)$, LoRA minimizes trainable parameters, lowering the barrier to injecting domain-specific knowledge into foundation models. 
\begin{figure}[ht]
  \centering
  \includegraphics[width=1.0\columnwidth]{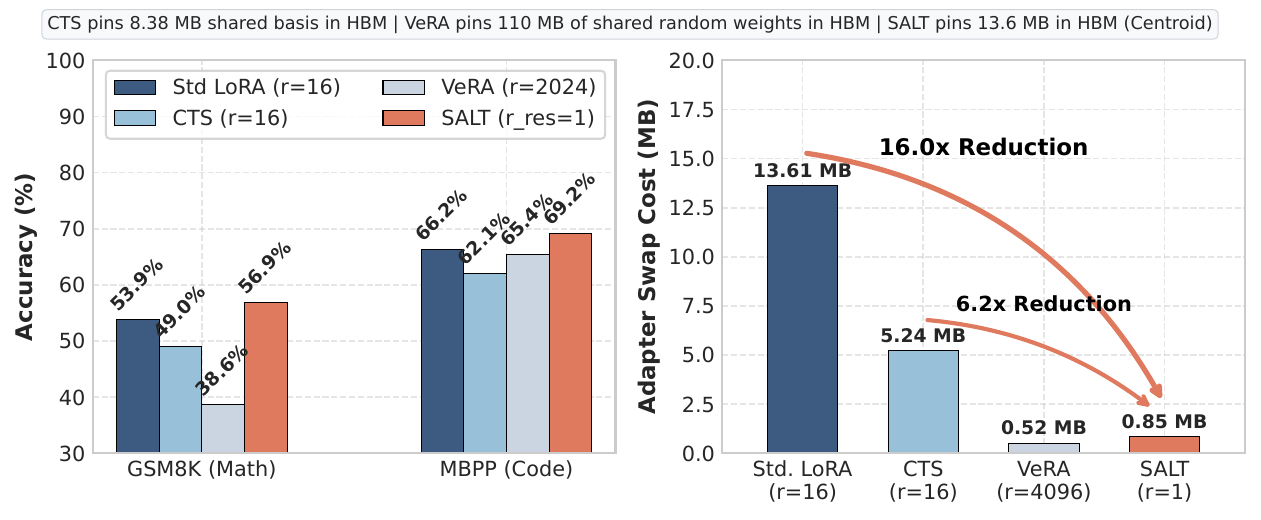} 
  \caption{\name\ reduces the PCIe bandwidth and VRAM usage by up to 16x while maintaining the performance.}
  \label{fig:performance_memory_tradeoff}
  \vspace{-3 mm}
\end{figure}
This tuning efficiency has driven the rapid growth of personalized LLM assistants and specialized agents~\cite{kong2024customizing,zhang2024personalization,li2026embodied}. Instead of deploying independent models, these specialized models are instantiated as distinct LoRA adapters attached to a shared foundation model. This structural shift has driven modern deployment toward \textit{LLM-as-a-Service (LLMaaS)} architectures~\citep{sheng2023s}, where centralized platforms (e.g., Amazon Bedrock~\cite{mohammed2026accelerate}) host thousands of custom, tenant-specific adapters simultaneously. To improve serving efficiency, Punica~\citep{chen2024punica} proposed Segmented Gather Matrix-Vector (SGMV) kernels that enable batched execution of heterogeneous adapters alongside a single frozen base model. However, multi-tenant LoRA serving remains bottlenecked by two hardware constraints: the Host-to-Device (H2D) PCIe swapping of adapter weights~\cite{toppings}, and the GPU VRAM required to host adapter weights alongside the KV-cache~\cite{sheng2023s}. This dual bottleneck creates a fundamental trade-off: higher rank adapters typically offer better downstream task performance but saturate VRAM capacity and PCIe bandwidth, whereas compressing these models to ultra-low ranks ($r \le 2$) reduces memory constraints but introduces drastic performance degradation.

Recent work has attempted to resolve those bottlenecks through compressing the post-training adapter or  decomposing adapters into shared and user-specific components. However, they face several critical limitations that render them unsuitable for real-world deployment.

\textit{Post-hoc Compression:} "Compress then Serve"~\cite{gabrielsson2025compress} applies joint diagonalization across hundreds of pre-trained LoRA adapters to extract a shared basis, reconstructing individual tasks via task-specific vectors. This approach may raise data isolation concerns in multi-tenant deployments~\cite{asif2026information} and requires recomputing the shared basis when boarding new tenants.

\textit{Continual Updates and Multi-Tenant Interference:}
Methods like Share~\cite{kaushik2026shared} incrementally update a shared subspace as each new user's task data arrives. This is structurally incompatible with multi-tenant deployment as each new user's data modifies the shared subspace, silently shifting the representations of all existing users. 

\textit{Frozen Shared Bases and Geometric Misalignment:}
Methods like VeRA~\citep{kopiczko2024vera} and  COLA~\citep{xia2024chain}) which freeze a random, task-agnostic foundation provide no structural shortcut for downstream learning, forcing highly constrained scaling vectors to compensate for a non-semantic basis, which severely bottlenecks expressivity and provides no mechanism to align independent adapters into a shared geometric basis.

\xiang{To bridge these gaps, we propose \textbf{S}ubspace-\textbf{A}ligned \textbf{L}oRA \textbf{T}raining (\textbf{SALT}), a serving-efficient fine-tuning framework designed explicitly for multi-tenant serving architectures. Unlike existing works, which require tenants to independently learn domain logic from scratch, \name\ intentionally shifts this heavy lifting to cloud providers. Our framework decomposes the LoRA weight update into two distinct components: a high-capacity, domain-shared centroid ($\bar{W}$) that aligns diverse intra-domain task adapters into a unified geometric basis, and an ultra-low-rank residual ($\delta_i$) that captures task-specific distribution. To make this completely seamless for the user, SALT employs a novel activation profiling mechanism that automatically routes unlabelled user data to the optimal domain centroid for fine-tuning. Cloud providers are highly incentivized to absorb this one-time training cost of those centroids: by anchoring users to shared foundations, tenants achieve higher rank performance using only ultra-low-rank residuals. During inference, the domain centroid (i.e., $r=16$) is permanently pinned in GPU memory as a shared anchor, and the serving engine only swaps the ultra-low rank residuals (e.g., $r=1$). \name\ not only alleviates the PCIe bandwidth bottleneck but also increases static adapter density, enabling serving engines to host vastly more concurrent tenants within the same hardware constraints, which significantly boosts the serving throughput.

Our main contributions are summarized as follows:}
\begin{itemize}
\item \xiang{To overcome the geometric divergence of independently trained adapters despite their shared domain knowledge, we introduce a matrix cosine regularizer that proactively aligns task subspaces into a unified basis.}
\item We propose \name, a serving-aware fine-tuning framework that decouples representational capacity from physical memory costs. By anchoring ultra-low-rank task residuals to a pinned, domain-shared centroid, \name\ recovers higher capacity adapter performance while minimizing the PCIe swapping footprint and GPU VRAM usage.
\item Across varying LLM scales, \name\ largely recovers higher rank performance using only $r=1$ task residuals. Furthermore, its up to 16x memory reduction boosts serving throughput by up to 51\% under PCIe bandwidth pressure and 28\% under GPU VRAM constraints with the Llama-3.2-3B model.
\end{itemize}
\section{Related Work}
\label{sec:related_work}

\begin{figure*}[h]
  \centering
  \includegraphics[width=1.0\textwidth]{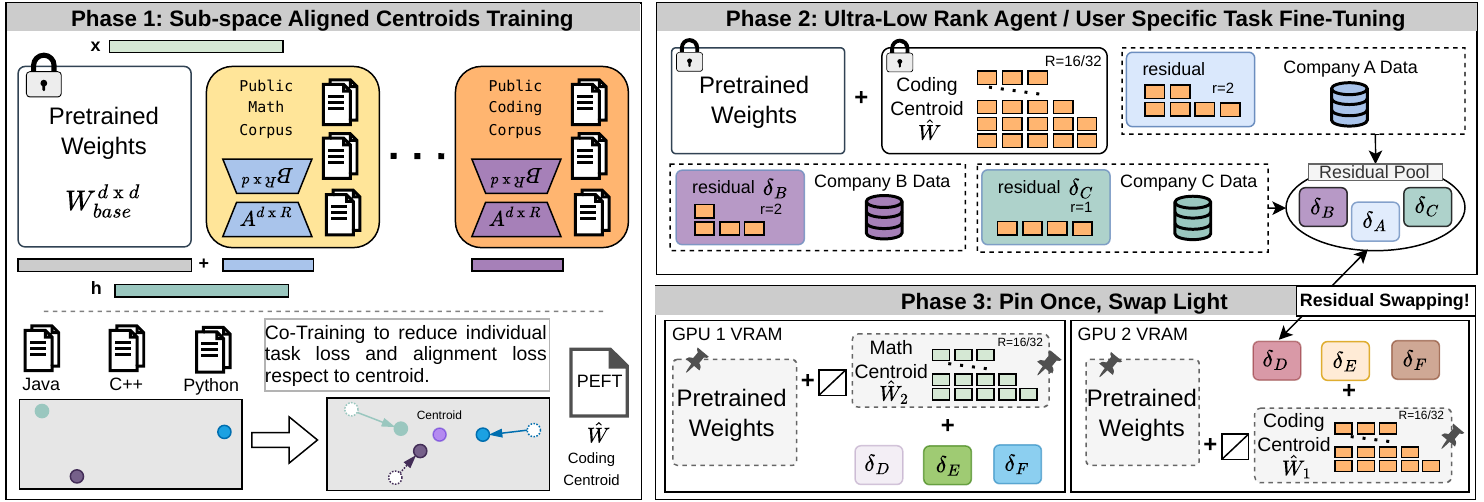} 
  \caption{The three phases of sub-space aligned fine-tuning for efficient LoRA serving. Phase 1: Training sub-space align centroids for different domains (i.e., Coding) with public corpus. Phase 2: Training ultra-low rank adapters using the domain centroids. Phase 3: Efficient multi-tenant serving with domain centroids and task-specific residuals.}
  \label{fig:wide_figure}
  \vspace{-3 mm}
\end{figure*}

\subsection{Multi-Tenant LoRA Serving}
Recent systems (i.e., Punica, SLoRA) optimize serving massive LoRA collections via custom kernels (Segmented Gather Matrix-Vector (SGMV) kernel), memory management or request co-scheduling~\citep{sheng2023s, chen2024punica, dlora}. Furthermore, CLIMB~\cite{zhang2026climb} has formalized the "LoRA residency cliff", demonstrating how rapid H2D swapping causes system-wide congestion collapse when KV-cache demands peak. Contrary to those optimizing at the infrastructure layer, \name\ operates at the algorithmic layer to produce ultra-low-rank ($r\le2$) residuals natively compatible with any SGMV-based engine, drastically reducing the payload these systems must manage.

\subsection{Shared Subspace and Adapter Compression}

Several works construct shared bases or manage adapter geometry to reduce multi-tenant memory overhead and mitigate interference. At initialization, methods like PiSSA~\cite{meng2024pissa} extract principal components directly from the base model via SVD to form an initial subspace. Post-training, "Compress then Serve"~\citep{gabrielsson2025compress} applies joint diagonalization to the pre-trained adapter to extract a shared orthonormal basis.
Similarly, EigenLoRAx~\cite{kaushik2025eigenlorax} extracts a principal subspace via SVD across a pre-trained adapter repository. However, this paradigm involves three practical trade-offs: (1) it requires pre-trained high-rank (e.g., $r=16$) adapters prior to compression; (2) the shared basis is derived directly from private task adapters, which may raise data isolation concerns in multi-tenant deployments~\cite{asif2026information}; and (3) adding or removing a user requires recomputing the shared basis, which grows costly as the adapter pool scales. Incremental approaches like Share~\citep{kaushik2026shared} and continual learning frameworks like LoDA~\citep{he2026task} similarly entangle representations and destroy isolation by continually updating shared principal factors with new user data. \name\ guarantees strict data isolation by training the shared centroid exclusively on provider-owned public data, and private data only updates the isolated task residual.

\subsection{Hierarchical and Residual Adaptation}
Decomposing adapters into shared and task-specific components is common in multi-task and continual learning. HydraLoRA~\citep{tian2024hydralora} employs a single shared $A$ matrix alongside multiple task-specific $B$ matrices via MoE routing, while  LoraHub~\cite{huang2023lorahub} compose independent LoRAs dynamically via few-shot inference. Shifting to sequential adaptation, LiLoRA~\cite{che2026lora} mitigates catastrophic forgetting in continual learning by maintaining a sequentially updated shared LoRA module, constrained by a cosine stability loss. However, these shared modules are continuously updated by diverse task data, which can entangle user representations and complicate the data isolation typically required for multi-tenant LLMaaS. Conversely, methods that use a frozen, randomly initialized base to ensure isolation (e.g., VeRA~\cite{kopiczko2024vera}) lack a semantically optimized multi-tenant anchor. COLA~\citep{xia2024chain} introduces additive residuals for single-task expansion but provides no mechanism to geometrically align independent adapters. Finally, methods like Align-LoRA~\cite{liu2025align} align low-dimensional representations produced by the shared LoRA projection via KL-divergence, which does not guarantee alignment of the underlying weight updates that our serving mechanism directly composes. \name\ bridges these gaps through an explicit training-time alignment that geometrically constrains task adapters to a public-data-trained centroid, enabling highly expressive rank-1 residuals natively suited for LLMaaS.

\section{Methods}\label{sec:methods}

\name\ introduces a serving-aware fine-tuning framework designed to break the memory bandwidth and footprint bottlenecks in the multi-tenant LoRA serving systems by decoupling the representational capacity of an adapter from its physical H2D transfer and GPU VRAM footprint.

\subsection{The Centroid-Residual Decomposition}
In a standard multi-tenant setting, the serving engine hosts a collection of $M$ distinct adapters, denoted as $\{\Delta W_i\}_{i=1}^M$, where each adapter is fine-tuned on its corresponding task dataset $\mathcal{D}_i$. A standard LoRA parameterizes the task-specific weight update as $\Delta W_i = B_i A_i$. During inference, dynamically swapping hundreds of these standard higher rank matrices (i.e., $r\ge16$) saturates PCIe bandwidth. Conversely, permanently pinning them in HBM severely exacerbates KV-cache starvation, bottlenecking concurrent tenancy.
To resolve this, \name\ structurally decomposes the weight update into two distinct components:
$$\Delta W_i = \bar{W} + \delta_i$$
\begin{itemize}
    \item \textbf{The Shared Centroid ($\bar{W}$):} A high-capacity, domain-specific, task-agnostic centroid (e.g., $r=16$) that is pre-loaded and pinned in the GPU's VRAM.
    \item \textbf{The Task Residual ($\delta_i$):} An ultra-low-rank dynamic adapter ($\delta_i = B'_i A'_i$, where $r \le 2$) that encapsulates task-specific knowledge. Only this minimal residual is dynamically swapped during multi-tenant inference.
\end{itemize}

\subsubsection{Phase 1: Joint Subspace Alignment}
A naive post-hoc averaging of independent adapters to establish $\bar{W}$ fails because isolated adapters converge to distinct, unaligned local minima. While joint diagonalization bypasses this, it violates multi-tenant data isolation and renders dynamic scaling intractable, as onboarding a new tenant requires recomputing the entire global basis.
To force in-domain task adapters to map their learned features into a unified, geometrically aligned subspace, in Phase 1 training, \name\ optimizes the individual task adapters ($\Delta W_i$) and the domain centroid ($\bar{W}$) jointly. Crucially, the centroid is not merely a passive geometric anchor; it is actively optimized on the multi-task data distribution. The joint objective is formulated as:
\begin{equation}
\begin{split}
\min_{\{\Delta W_i\}, \bar{W}} \frac{1}{M} \sum_{i=1}^M w_i\Big[ & \mathcal{L}_{\text{task}}(W_{base}+\Delta W_i; \mathcal{D}_i) + \\ \mathcal{L}_{\text{task}}(W_{base}+\bar{W}; \mathcal{D}_i)
& + \lambda \mathcal{L}_{\text{align}}(W_{base}+\Delta W_i, \bar{W}) \Big]
\end{split}
\end{equation}
where $\lambda$ controls the alignment penalty and $w_i \propto |\mathcal{D}_i|$ reweights each task by its relative dataset size, preventing smaller, more frequently cycled datasets from being overrepresented. To guarantee strict privacy, Phase 1 is executed by the cloud provider using only public, domain-specific datasets; user-private data is strictly reserved for Phase 2 fine-tuning.
We implement $\mathcal{L}_{\text{align}}$ as an epsilon-stabilized matrix cosine regularizer:
\begin{equation}
\mathcal{L}_{\text{align}}(\Delta W_i, \bar{W}) = 1 - \frac{\text{Tr}(\Delta W_i^T \bar{W})}{\|\Delta W_i\|_F \|\bar{W}\|_F + \epsilon}
\end{equation}

The constant ($\epsilon = 10^{-8}$) prevents numerical instability and division-by-zero artifacts at initialization ($\bar{W} \approx 0$).

\subsubsection{Phase 2: Ultra-Low Rank Residual Tuning}
Once the geometrically aligned centroid ($\bar{W}$) is established, Phase 2 adapts the model to private, user-specific data. To guarantee strict multi-tenant data isolation, the optimization is exclusively confined to the ultra-low-rank task residual $\delta_i$:
\begin{equation}
\min_{\delta_i} \mathcal{L}_{\text{task}}(W_{\text{base}} + \bar{W} + \delta_i; \mathcal{D}_{\text{private}}^{(i)})
\end{equation}

Constraining this frozen-anchor optimization to ultra-low ranks ($r \le 2$) is not merely a systems-level memory optimization; it also acts as a structural regularizer. Because the constrained residual lacks the parameter capacity to model the target distribution independently, it acts as a geometric correction vector. It is forced to utilize the pre-aligned centroid as a structural shortcut---achieving constructive alignment with the generalized features. This dynamic allows \name\ to successfully recover higher-rank standard LoRA performance while drastically reducing the dynamic H2D PCIe swapping footprint.

\begin{table*}[ht]
\centering
\fontsize{9pt}{\baselineskip}\selectfont 
\setlength{\tabcolsep}{1mm} 
\begin{tabular}{@{}llcc@{\hspace{0.3em}}cccc@{\hspace{0.3em}}ccc@{\hspace{0.3em}}c@{}}
\toprule
 & & \multicolumn{2}{c}{\textbf{Memory (MB)}} & \multicolumn{4}{c}{\textbf{Math Reasoning Tasks}} & \multicolumn{3}{c}{\textbf{Coding Tasks}} & \\
\cmidrule(lr){3-4} \cmidrule(lr){5-8} \cmidrule(lr){9-12}
\textbf{Method} & \textbf{Rank} & \textbf{Pin} & \textbf{Swap} & \textbf{GSM8K}$^\ddagger$ & \textbf{SVAMP}$^\ddagger$ & \textbf{Multiarith}$^\dagger$ & \textbf{AQuA}$^\dagger$ & \textbf{MBPP}$^\ddagger$ & \textbf{SPIDER}$^\ddagger$ & \textbf{APPs}$^\dagger$  & \textbf{HumanEval} \\
\midrule
Base & NA & -- & -- & 3.93 & 6.67 & 6.90 & 24.51 & 4.50 & 0.00 & 8.75 & 18.29 \\
\midrule
\multirow{5}{*}{LoRA} 
 & $r=1$ & 0.00 & 0.85 & 49.67 & 63.33 & 11.03 & \textbf{34.80} & 15.75 & 30.34 & 16.0 & 25.58 \\
 & $r=2$ & 0.00 & 1.70 & 50.80 & 63.75 & 18.62 & \textbf{36.76} & 24.50 & 29.50 & 12.25 & 12.40 \\
 & $r=4$ & 0.00 & 3.40 & 53.13 & 65.00 & 27.58 & \textbf{38.24} & 25.50 & 9.47 & \textbf{21.25} & 21.71 \\
 & $r=8$ & 0.00 & 6.81 & 53.70 & 64.17 & 26.90 & 35.29 & 3.00 & 36.21 & 16.75 & 24.81 \\
 & $r=16$ & 0.00 & 13.61 & 53.88 & 66.25 & 30.35 & 34.80 & 26.75 & 21.34 & 18.75 & 21.71 \\
\midrule
\multirow{3}{*}{\begin{tabular}{@{}l@{}}LoRA\\ (Merged \\Dataset)\end{tabular}}
 & $r=1$ & 0.00 & 0.85 & 48.08 & 65.0 & -- & -- & \textbf{29.75} & 50.95 & -- & -- \\
 & $r=2$ & 0.00 & 1.70 & 51.63 & 67.08 & -- & -- & \textbf{28.75} & 52.63  & -- & -- \\
 & $r=4$ & 0.00 & 3.40 & 51.63 & 67.91 & -- & -- & 26.75 & 52.52 & -- & -- \\
\midrule
\multirow{3}{*}{VeRA} 
 & $r=1k$ & 27.26 & 0.26 & 40.51 & 60.83 & 36.55 & 32.35 & 24.25 & 21.22 & 17.0 & 18.60 \\
 & $r=2k$ & 54.52 & 0.52 & 38.63 & 65.42 & 31.03 & 34.80 & 24.75 & 24.34 & 17.5 & 19.37 \\
 & $r=4k$ & 109.05 & 1.05 & 40.79 & 60.83 & 32.41 & 27.45 & 26.50 & 31.53 & 17.75 & 20.16 \\
\midrule
\multirow{5}{*}{\begin{tabular}{@{}l@{}}CTS \\ (JD-Full)\end{tabular}} 
 & $r=1$ & 0.52 & 0.33 & 9.17 & 40.83 & 18.62  & 24.02 & 2.75 & 0.0 & 11.5 & 17.05 \\
 & $r=2$ & 1.05 & 0.66 & 10.20 & 66.25 & 27.59 & 23.04 & 14.25 & 0.12 & 14.75 & 13.17 \\
 & $r=4$ & 2.01 & 1.31 & 37.69 & 64.58 & 26.90 & 25.49 & 5.00 & 0.0 & 8.25 & 17.82 \\
 & $r=8$ & 4.19 & 2.62 & 47.05 & 63.33 & 28.28 & 34.31 & 2.50  & 0.12 & 7.75 & 17.05 \\
 & $r=16$ & 8.38 & 5.24 & 49.02 & 62.08 & 31.72 & 29.90 & 9.25 & 3.47 & 10.5 & 13.95 \\
\midrule
\multirow{3}{*}{\begin{tabular}{@{}l@{}}\textbf{\name{}} \\ \textbf{(Ours)}\end{tabular}} 
 & $r=1$ & 13.63 & 0.85 & \textbf{56.88} {\scriptsize (+7.2)} & \textbf{69.17} {\scriptsize (+5.8)} & \textbf{86.21} {\scriptsize (+73.9)} & 34.31 {\scriptsize (-0.4)} & 27.25 {\scriptsize (+11.5)} & \textbf{56.59} {\scriptsize (+26.3)} & \textbf{19.25} {\scriptsize (+3.2)} & \textbf{34.88} {\scriptsize (+9.3)} \\
 & $r=2$ & 13.63 & 1.70 & \textbf{57.81} {\scriptsize (+7.0)} & \textbf{69.17} {\scriptsize (+5.4)} & \textbf{85.52} {\scriptsize (+66.4)} & 31.37 {\scriptsize (-5.3)} & 27.80 {\scriptsize (+3.3)} & \textbf{56.35} {\scriptsize (+26.9)} & \textbf{17.75} {\scriptsize (+5.5)} & \textbf{34.11} {\scriptsize (+21.7)} \\
 & $r=4$ & 13.63 & 3.40 & \textbf{53.23} {\scriptsize (+0.1)}  & \textbf{68.75} {\scriptsize (+3.7)} & \textbf{86.21} {\scriptsize (+59.6)} & 31.86 {\scriptsize (-6.3)} & \textbf{27.25} {\scriptsize (+1.8)} & \textbf{56.47} {\scriptsize (+47.0)} & 18.25 {\scriptsize (-3.0)} & \textbf{35.66} {\scriptsize (+13.9)} \\
\bottomrule
\end{tabular}

\raggedright
$^\ddagger$ Datasets utilized for Phase 1 centroid training and Phase 2 residual tuning. $^\dagger$ Datasets for Phase 2 residual tuning only. LoRA merged dataset refers to a single standard LoRA adapter trained on the concatenation of $^\ddagger$ datasets. The numbers in parenthesis are improvement compared to the standard LoRA at the same rank.
\caption{Peak task performance and memory footprint for Mistral-7B-v0.3. \name\ at r=1 largely recovers the performance of higher rank standard LoRA with significantly reduced memory footprint.}
\label{tbl:mistral_table}
\end{table*}

\subsection{Automated Centroid Routing}
When a user submits an unlabelled dataset $\mathcal{D}$ for Phase 2 fine-tuning, the system must automatically select the optimal centroid from the Phase 1 centroid library $\mathcal{C} = \{\bar{W}_1, \dots, \bar{W}_K\}$, or fallback if the user data is entirely out-of-distribution (OOD). We achieve this robustly using a two-stage activation profiling algorithm. 

Using a minimal subset of $\mathcal{D}$, we compute the centroid k-induced displacement norm $N_k = \| h_k - h_{\text{base}} \|_2$ for each centroid, where $h$ is the final hidden state. To guarantee length invariance, this metric is token-normalized. The system evaluates these novel activations against baseline distributions ($\mu, \sigma$). These baselines are established offline at the end of Phase 1 by passing each centroid's own in-domain validation set through the model and recording the empirical mean and standard deviation of its activations. Using these pre-calibrated statistics, the system applies a two-stage filter:

\textbf{Stage 1: OOD Rejection.} Building on representation-based OOD detection~\cite{ren2021simple,sun2021react}, we compute the deviation $Z_{N,k} = \frac{|N_k - \mu_{N,k}|}{\sigma_{N,k}}$. If the minimum deviation across all centroids exceeds a strict threshold ($\min_k Z_{N,k} > \tau_N$, e.g., $\tau_N = 2.0$ for 95\% confidence interval (CI)), the data lacks geometric overlap with the available library. The system flags it as OOD and falls back to standard LoRA training.

\textbf{Stage 2: Expert Routing.} For in-domain tasks, absolute norms naturally fluctuate based on prompt structure and vocabulary complexity. To isolate true domain alignment, we compute a dominance ratio $R_k$, comparing a centroid's activation against the average of all others. The optimal centroid $k^*$ is selected by minimizing the ratio deviation $Z_{R,k}$:
\begin{equation}
R_k = \frac{N_k}{\frac{1}{K-1} \sum_{j \neq k} N_j}, \ \quad k^* = \arg\min_k \frac{|R_k - \mu_{R,k}|}{\sigma_{R,k}}
\end{equation}
If the minimum deviation still exceeds the routing threshold ($Z_{R,k^*} > \tau_R$, e.g., $\tau_R = 2.0$), the task is considered a blended or ambiguous domain and sent to the fallback. Otherwise, the system automatically routes the user's dataset to $\bar{W}_{k^*}$ as the anchor for Phase 2 task residual tuning.

\subsection{Multi-Tenant Serving} \label{subsec:multi-serving}
During multi-tenant serving, the final weight matrix for a user request $i$ is instantiated as:
$$W_{final} = W_{base} + \gamma \overline{W} + \delta_i$$
\xiang{While the task residual $\delta_i$ is optimized against the unscaled centroid during Phase 2, directly composing dense, multi-task representations at inference time can cause interference. Following established principles in task arithmetic \cite{ilharco2023editing}, we introduce an inference-time scaling coefficient $\gamma$ to dampen multi-task interference ($\gamma < 1.0$) or amplify target capabilities ($\gamma > 1.0$). 
To avoid the overhead of dynamically composing continuous $\gamma$ values, SALT discretizes $\gamma$ into predefined bins (e.g., $\gamma \in \{0.6, 0.8, 1.0\}$). The serving frontend groups incoming requests by their assigned ($\overline{W}, \gamma$) pair. For high-frequency domain centroids, providers can fuse the scaled centroid directly into the base weights at node startup. During inference, these nodes only dynamically swap the ultra-low-rank residuals ($\delta_i$). For the long tail of diverse requests, the scaled centroid and the residual can simply be concatenated offline into a single standard adapter (e.g., $r=17$). This allows the serving engine to seamlessly handle rare tasks using standard heterogeneous batching.}

\section{Experiments}

\subsection{Evaluation Setup}\label{subsec:exp_setup}
\noindent \textbf{Datasets}:
We evaluate \name{} across two domains — mathematical reasoning and coding — using three base model families of varying scale (Llama-3.2-3B, Mistral-7B-v0.3, Pythia-12B). Math datasets include GSM8K~\cite{cobbe2021gsm8k}, SVAMP~\cite{patel-etal-2021-nlp}, MultiArith~\cite{roy-roth-2015-solving}, and AQuA~\cite{ling-etal-2017-program}; coding tasks include MBPP~\cite{austin2021program}, SPIDER~\cite{yu-etal-2018-spider}, APPS~\cite{hendrycksapps2021}, and HumanEval~\cite{chen2021evaluating}. Dataset details are listed in Appendix~\ref{subsec:implementation_details}.

\noindent \textbf{Evaluation Metrics}: We report exact-match accuracy for math tasks, functional correctness (unit tests) for MBPP/HumanEval/Apps, and structural exact set match for Spider. All \name{} results use the scaling factor $\gamma$ selected via a small hold-out validation set, disjoint from the test sets in Table~\ref{tbl:mistral_table}.

\noindent \textbf{Domain Centroids \& Residuals}: In Phase 1, the Math centroid is trained on GSM8K and SVAMP, and the Code centroid on MBPP and Spider. For Phase 2, we evaluate three settings: (1) \textit{Seen In-Domain}:  residuals tuned on the Phase 1 datasets ($\ddagger$ in Table~\ref{tbl:mistral_table}); (2) \textit{Unseen In-Domain}: residuals tuned on held-out datasets (MultiArith, AQuA, APPs; $\dagger$ in Table~\ref{tbl:mistral_table}) to test generalization; (3) \textit{Zero-Shot Transfer}: the MBPP-tuned residual is evaluated directly on HumanEval without further training, testing whether the shared centroid enables generalization to a related, unseen task.

\noindent \textbf{Hardware}: All throughput and latency experiments are conducted on NVIDIA H100 GPUs.


\subsection{Baselines}
We compare against LoRA (isolated, uncompressed task-specific fine-tuning), VeRA (a frozen, randomly initialized shared projection with trainable scaling vectors), and Compress then Serve (CTS) (post-training joint diagonalization of independent adapters into a shared basis). We restrict comparison to VeRA and CTS as they are the primary prior methods that, like \name{}, produce a compressed per-tenant serving artifact; other related methods target multi-task or continual-learning objectives without an explicit serving-efficiency mechanism are discussed in Section~\ref{sec:related_work}.

\subsection{Task Residual Adapter Performance }
\begin{figure}[ht]
  \centering
  \includegraphics[width=1.0\columnwidth]{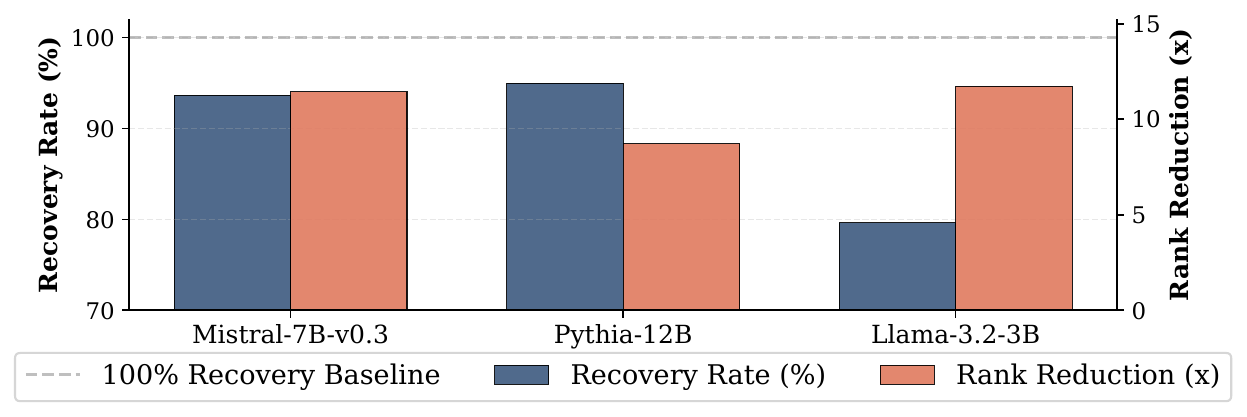} 
  \caption{Average task performance recovery and adapter rank reduction for three evaluated models.}
  \label{fig:recovery_all}
\end{figure}

\xiang{As shown in Table~\ref{tbl:mistral_table}, \name\ largely outperforms standard LoRA and state-of-the-art compression baselines in downstream task performance while utilizing an ultra-low-rank residual. VeRA's reported rank (1024-4096) is the dimensionality of the frozen random matrix, not a trainable adapter rank; since only diagonal scaling vectors are learned, a large basis is needed for competitive expressivity. Although high-capacity standard LoRA maintains a slight edge on a few specific datasets (e.g., AQuA), with a dynamic footprint of just 0.85 MB ($r=1$), \name\ consistently exceeds both standard low-rank configurations and the state-of-the-art compression baselines. This suggests that anchoring residuals to a pre-aligned centroid decouples downstream task performance from multi-tenant PCIe bottlenecks.
Moreover, \name\ avoids the severe accuracy degradation and rank-sensitivity seen in baselines: standard LoRA and CTS exhibit volatile rank-to-rank fluctuations (e.g., LoRA drops to 3.00\% at $r=8$ on MBPP despite scoring 25.50\% at $r=4$), whereas \name\ remains stable across $r \in \{1,2,4\}$. This suggests that anchoring residuals to a shared centroid helps regularize against the instability that unconstrained low-rank adapters can exhibit.
Figure~\ref{fig:recovery_all} shows these gains are architecture-agnostic. Across Mistral-7B-v0.3, Pythia-12B, and Llama-3.2-3B, \name\ consistently achieves an average of 80–95\% downstream task performance recovery  while delivering 9x to 11.5x reductions in dynamic adapter rank. Detailed comparison results for \name\ and baselines on each model can be found in the Appendix~\ref{subsec:full_results}.}

\subsection{Serving Efficiency and Throughput}
\begin{figure}[h]
  \centering
  \includegraphics[width=1.0\columnwidth]{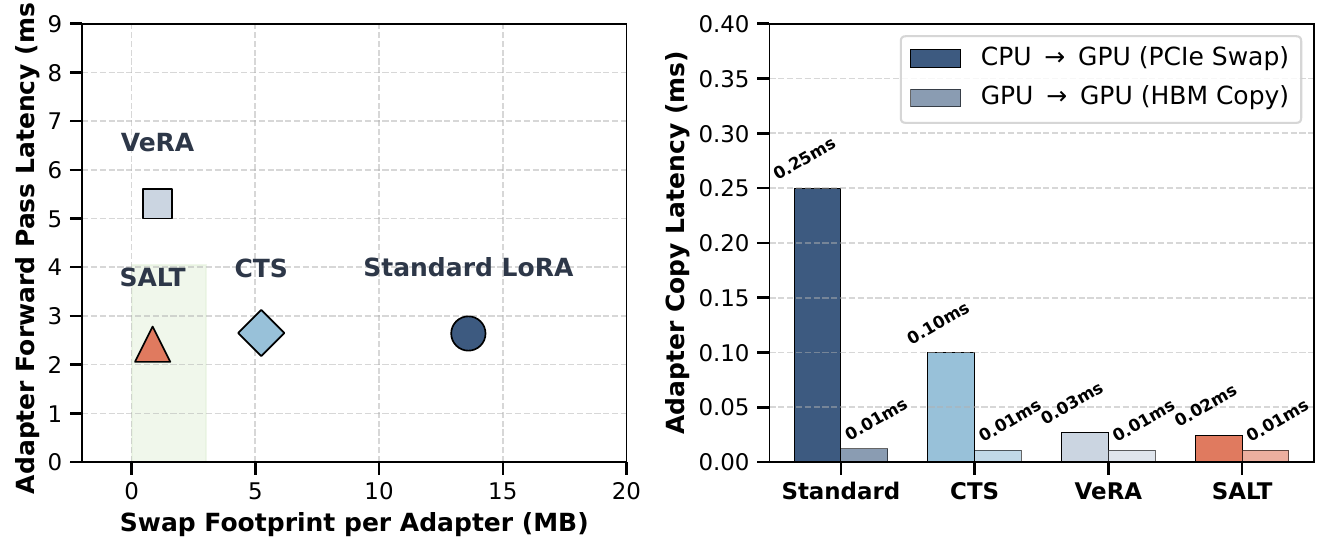} 
  \caption{Mistral: (Left) Swap footprint versus adapter forward pass latency. (Right) Latency breakdown for PCIe swap and HBM copy operations across different baselines.}
  \label{fig:micro_bench_mistral}
\end{figure}

\begin{table}[ht]
\centering
\fontsize{9pt}{\baselineskip}\selectfont 
\setlength{\tabcolsep}{2mm} 
\begin{tabular}{@{}lrrrr@{}}
\toprule
 & \multicolumn{2}{c}{\textbf{PCIe}} & \multicolumn{2}{c}{\textbf{VRAM}} \\
\cmidrule(lr){2-3} \cmidrule(l){4-5}
\textbf{Adapters} & \textbf{vLLM} & \textbf{vLLM+SALT} & \textbf{vLLM} & \textbf{vLLM+SALT} \\
\midrule
128 & 98.38  & 107.77\,\scriptsize{(+9.5\%)}  & 77.82  & 82.50\,\scriptsize{(+6.0\%)} \\
256 & 80.79  & 97.47\,\scriptsize{(+20.6\%)}  & 45.82  & 49.91\,\scriptsize{(+8.9\%)} \\
512 & 43.42  & 54.46\,\scriptsize{(+25.4\%)}  & 32.59  & 38.41\,\scriptsize{(+17.9\%)} \\
\bottomrule
\end{tabular}
\caption{Throughput for Mistral-7B-v0.3 (req/s).}
\label{tab:throughput_results_mistral}
\end{table}
To demonstrate serving efficiency, we evaluate \name\ with vLLM and scaling the number of concurrent adapters ($N$) up to 512. We assess throughput across two extreme memory constraints - PCIe bandwidth and HBM capacity pressure:
\begin{enumerate}
    \item Constraining the adapter cache to force continuous H2D swapping, \name\ yields +25.0\% throughput over standard LoRA at $N=512$.
    \item When all $N$ residual adapters instead reside in GPU VRAM, Standard LoRA's large footprint rapidly consumes the KV-cache, forcing the scheduler to fragment batch sizes and preempt requests to avoid out-of-memory errors. \name's $16\times$ footprint reduction avoids this, sustaining larger continuous batches and improving peak throughput by up to 17.9\% at $N=512$.
\end{enumerate}
Since \name\ relies on standard additive matrix operations, these throughput gains are achieved without custom kernels, ensuring compatibility with production engines. We omit VeRA and CTS from this end-to-end evaluation as their specialized matrices require non-standard serving pipelines and cannot be directly used with vLLM. Instead, Figure~\ref{fig:micro_bench_mistral} provides micro-benchmarks against all baselines. \name\ achieves the lowest forward-pass latency while requiring an order-of-magnitude smaller swap footprint. Moreover, \name\ matches VeRA's minimal copy latency and substantially outperforms Standard LoRA and CTS.

\begin{table}[ht]
\centering
\fontsize{9pt}{\baselineskip}\selectfont 
\setlength{\tabcolsep}{2.5mm} 
\begin{tabular}{@{}lcccc@{}}
\toprule
 & & \multicolumn{2}{c}{\textbf{Ratio Dev. ($Z_R$)}} & \\
\cmidrule(lr){3-4}
\textbf{Task} & \textbf{$Z_N$} & \textbf{Math} & \textbf{Code} & \textbf{Routed To} \\
\midrule
\texttt{AQuA}       & 1.32 & \textbf{0.94} & 2.75 & Math \\
\texttt{MultiArith} & 0.39 & \textbf{0.11} & 3.40 & Math \\
\texttt{CodeSearchNet} & 1.75 & 3.55 & \textbf{0.49} & Code \\
\texttt{APPs} & 1.48 & 3.12 & \textbf{0.39} & Code \\
\texttt{xnli/german} & \textbf{2.28} & 1.65 & 2.12 & OOD \\
\texttt{xnli/spanish} & \textbf{2.32} & 1.69 & 2.19 & OOD \\
\bottomrule
\end{tabular}
\caption{Zero-shot hybrid centroid routing validation on Mistral-7B-v0.3 ($\tau_N=2.0, \tau_R=2.0$, corresponding to standard 95\% CI threshold).}
\label{tab:zero_shot_routing}
\end{table}

\subsection{Subspace Alignment Experiment}
\begin{figure}[ht]
  \centering
  \includegraphics[width=1.0\columnwidth]{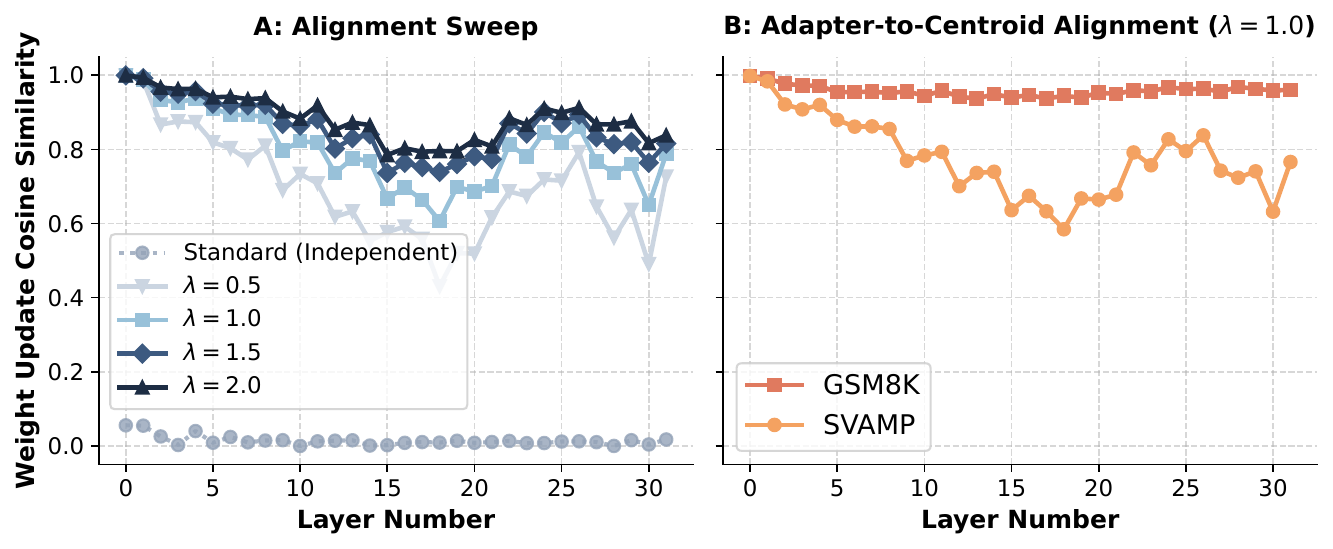} 
  \caption{Left: Cosine similarity between GSM8K and SVAMP weight updates across layers. Right: Cosine similarity of individual adapters to the shared centroid ($\lambda = 1.0$).}
  \label{fig:subspace_align}
\end{figure}

We validate Phase 1 subspace alignment by analyzing layer-wise cosine similarity of full adapter updates ($\Delta W = BA$) for GSM8K and SVAMP. As Figure~\ref{fig:subspace_align} (Left) shows, independently trained Standard LoRA updates for the two adapters exhibit near-zero similarity despite sharing domain semantics. However, under \name{}'s alignment penalty, as $\lambda$ increases, independently learned updates are forced to converge into a shared subspace. Moreover, this forces both adapters and the centroid into the same aligned subspace as shown in Figure~\ref{fig:subspace_align}(right).

\subsection{Phase 2 Centroid Routing}

We evaluate automated centroid routing on six datasets held out from Phase 1 training: AQuA and MultiArith (math), CodeSearchNet~\cite{husain2019codesearchnet} and APPs (code), and two out-of-distribution (OOD) language splits from XNLI~\cite{conneau2018xnli} (German, Spanish). Following our protocol, we sample $|\mathcal{D}_{\text{profile}}| = 100$ examples from each dataset and extract adapter-induced displacement norms against both centroids.
As shown in Table~\ref{tab:zero_shot_routing}, the system correctly isolates OOD data while routing in-domain tasks accurately. For both language tasks, the adapter activation norm deviates sharply from both centroids' expected ranges ($Z_N = 2.28$ and $2.32$, both $> \tau_N$), this flags the data as OOD and triggers a fallback to Standard LoRA training. The remaining four datasets pass the OOD check ($Z_N \le 2.0$) and are routed by minimum ratio deviation: AQuA and MultiArith align with the Math centroid ($Z_R = 0.94$ and $0.11$), while CodeSearchNet and APPs route to the Code centroid ($Z_R = 0.49$ and $0.39$). These results show \name\ can automate centroid selection for in-domain tasks while gracefully handling out-of-distribution data.

\section{Ablation Studies}
\subsection{Alignment Penalty $\lambda$ Sensitivity}
\begin{figure}[ht]
  \centering
  \includegraphics[width=1.0\columnwidth]{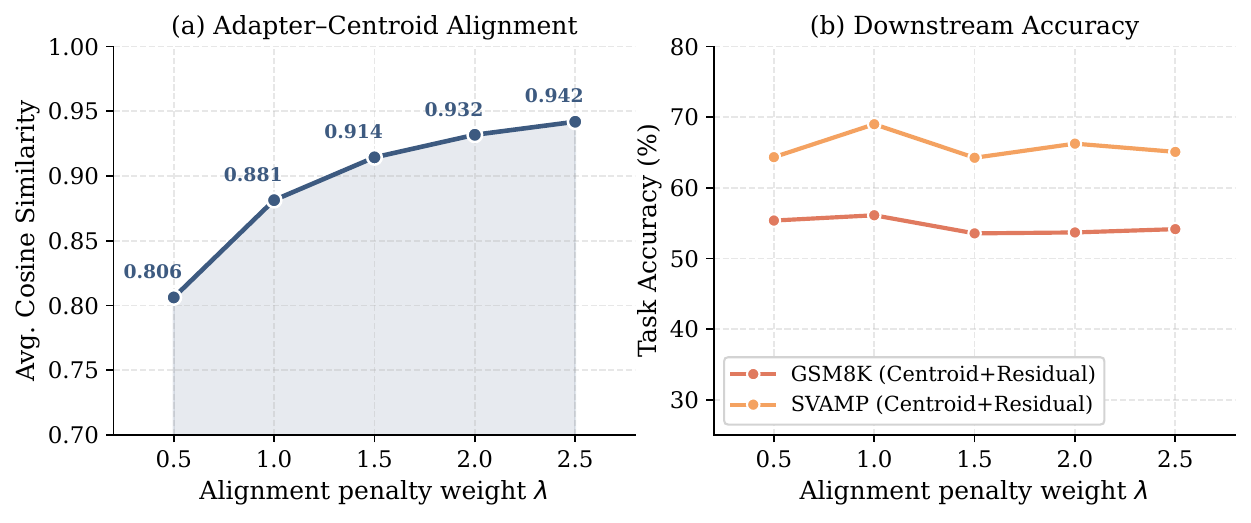} 
  \caption{Effect of alignment penalty $\lambda$.}
  \label{fig:lambda_ablation}
\end{figure}
We performed a sensitivity analysis on the alignment loss weight $\lambda$. We swept $\lambda$ from 0.5 to 2.5, and show the corresponding downstream task accuracy. As shown in Figure~\ref{fig:lambda_ablation}, increasing $\lambda$ monotonically improves the average adapter-centroid cosine similarity across all layers. Despite stronger regularization, average downstream accuracy within the domain remains stable, demonstrating that alignment incurs no significant performance penalty for in-domain tasks.

\subsection{Centroid Scale $\gamma$ Sensitivity}

\begin{table}[ht]
\centering
\fontsize{9pt}{\baselineskip}\selectfont 
\setlength{\tabcolsep}{9pt} 
\begin{tabular}{@{}lccccc@{}}
\toprule
 & \multicolumn{5}{c}{\textbf{Scale Factor ($\gamma$)}} \\
\cmidrule(l){2-6}
\textbf{Dataset}  & \textbf{0.4} & \textbf{0.6} & \textbf{0.8} & \textbf{1.0} & \textbf{1.2} \\
\midrule
GSM8K  & 53.04 & 58.46 & \textbf{58.65} & 55.29 & 48.83 \\
SVAMP  & 65.83 & \textbf{70.0} & 68.33 & \textbf{70.0} & 66.25 \\
AQuA   & 26.96 & \textbf{28.43} & 27.94 & \textbf{28.43} & 25.98 \\
\bottomrule
\end{tabular}
\caption{Effect of centroid output scale factor ($\gamma$) on downstream performance (\%) for Mistral-7B-v0.3.}
\label{tab:centroid_scale_sweep}
\end{table}

As shown in Table~\ref{tab:centroid_scale_sweep}, downstream performance follows an inverted-U pattern across the scale factor $\gamma$, peaking in the $\gamma=0.6$--$0.8$ range and degrading at both extremes. Notably, performance across $\gamma=0.6$, $0.8$, and $1.0$ remains tightly clustered, suggesting the system is fairly insensitive to the exact scale factor within this middle range. This is practically useful for multi-tenant deployment: rather than fine-tuning $\gamma$, a provider can adopt few shared $\gamma$ values (e.g., 0.8) across domains with only a marginal accuracy trade-off. Performance only degrades meaningfully at extreme boundaries ($\gamma=0.4$ or $1.2$), indicating that precise fine-tuning is unnecessary as long as these boundary regions are avoided.

\begin{table}[ht]
\centering
\fontsize{9pt}{\baselineskip}\selectfont 
\setlength{\tabcolsep}{2mm} 
\begin{tabular}{@{}ccccc@{}}
\toprule
\textbf{Rank ($r_{\text{centroid}}$)} & \textbf{GSM8K} & \textbf{SVAMP} & \textbf{MultiArith} & \textbf{AQuA} \\
\midrule
8  & 55.47 & 68.33 & 85.51 & 32.35 \\
16 & 56.88 & 69.17 & 86.21 & 34.31 \\
32 & 55.29 & 67.50 & 89.65 & 30.39\\
\bottomrule
\end{tabular}
\caption{Phase 1 Centroid Rank ($r_{\text{centroid}}$) analysis.}
\label{tab:centroid_rank_ablation}
\end{table}


\subsection{Centroid Rank Sensitivity}
We evaluate the impact of Phase 1 centroid capacity on downstream performance. As shown in Table~\ref{tab:centroid_rank_ablation}, increasing rank from $r=8$ to $r=16$ improves all four datasets, but scaling further to $r=32$ degrades GSM8K, SVAMP, and AQuA, with only MultiArith continuing to improve (89.65\%). This indicates $r=16$ offers the best capacity--generalization balance for this domain. Since higher-rank centroids also incur additional pinned VRAM, this result demonstrates that a moderate centroid rank is both the most accurate and memory-efficient choice in our settings. Centroid training remains efficient across this range (Appendix~\ref{subsec:training_src}), so exploring rank during offline domain setup is inexpensive.

\subsection{Effect of Subspace-Aligned Centroid}\label{subsec:sub_t_test}
We isolate the effect of subspace-aligned training against a centroid trained on naive dataset concatenation. Table~\ref{tab:ablation_centroid} shows results at $r=2$; subspace alignment consistently outperforms the mixed-data. Across 9 paired configurations ($r \in \{1,2,4\}$ over GSM8K, SVAMP, and MBPP), the mean gain is $+4.38\%$ (paired $t$-test, $p=2.5\times10^{-5}$), confirming a statistically significant and robust improvement. The result and a detailed analysis on SPIDER can be found separately in Appendix~\ref{subsec:full_subspace_analysis}, as it exhibits high variance across ranks.
\begin{table}[ht]
\centering
\fontsize{9pt}{\baselineskip}\selectfont
\setlength{\tabcolsep}{2mm} 
\begin{tabular}{@{}lcccc@{}}
\toprule
 & \multicolumn{4}{c}{\textbf{Accuracy \%}} \\
\cmidrule(l){2-5}
\textbf{Centroid Method} & \textbf{GSM8K} & \textbf{SVAMP} & \textbf{MBPP} & \textbf{SPIDER} \\
\midrule
Concatenated & 52.94 & 63.75 & 22.5 & 12.47\\
Subspace-aligned & \textbf{57.81} & \textbf{69.17} & \textbf{27.8} & \textbf{56.35} \\
\bottomrule
\end{tabular}
\caption{Effect of subspace-aligned centroid training strategy ($r=2$) on downstream tasks For Mistral-7B-v0.3.}
\label{tab:ablation_centroid}
\end{table}


\section{Limitations}
While \name{} demonstrates significant serving efficiency, it introduces a structural bias that may penalize specific in-domain tasks. Because Phase 1 subspace alignment enforces a strict geometric intersection, tasks with orthogonal formats like multiple-choice evaluation within a predominantly direct-generation domain can experience quality degradation. In these cases, the highly constrained $r=1$ residual lacks the capacity to overcome the centroid's structural regularization, occasionally resulting in worse performance than standard LoRA like AQuA in Table~\ref{tbl:mistral_table}. Moreover, our evaluation is restricted to two domains (math and coding); as shown in Appendix~\ref{subsec:mix_domain}, combining unrelated domains into a single centroid may degrade in-domain accuracy, indicating that \name's benefits depend on the provider partitioning tasks into sufficiently coherent domains. Finally, extending \name{} to Mixture-of-Experts architectures remains open: MoE's sparse, disjoint routing paths make establishing a unified geometric subspace across experts mathematically complex without additional memory cost. 
\section{Conclusion}
We introduced \name, a serving-aware fine-tuning framework that anchors ultra-low-rank residuals to subspace-aligned domain centroids to recover higher-rank task performance while reducing per-adapter memory by up to 16x, ultimately boosting serving throughput by up to 51\% under PCIe bandwidth pressure and 28\% under GPU VRAM constraints.

\bibliography{aaai2027}

\clearpage

\clearpage
\appendix
\section{Appendix}
\subsection{Experiment Setup Details}\label{subsec:implementation_details}
\begin{table}[ht]
\centering
\fontsize{9pt}{\baselineskip}\selectfont
\setlength{\tabcolsep}{6mm}
\begin{tabular}{@{}lccc@{}}
\toprule
\textbf{Dataset} & \textbf{Train} & \textbf{Valid} & \textbf{Test} \\
\midrule
GSM8K$^{\ddagger}$    & 7473 & 250 & 1069 \\
SVAMP$^{\ddagger}$    & 700     & 60  & 240  \\
MultiArith$^{\dagger}$ & 420    & 35  & 145  \\
AQuA$^{\dagger}$      & 5000 & 50  & 204  \\
\midrule
MBPP$^{\ddagger}$     & 374     & 100 & 400  \\
SPIDER$^{\ddagger}$   & 7000 & 200 & 1034 \\
APPs$^{\dagger}$      & 2000 & 100 & 400  \\
HumanEval$^{\dagger\dagger}$ & --  & --  & 164  \\
\bottomrule
\end{tabular}
\caption{Dataset details. $\ddagger$: used for Phase 1 centroid training and Phase 2 residual tuning. $\dagger$: used for Phase 2 residual tuning only (Unseen In-Domain). $\dagger\dagger$: used exclusively for zero-shot evaluation (Section~\ref{subsec:exp_setup}); no training split is used.}
\label{tab:dataset_stats}
\end{table}

Table~\ref{tab:dataset_stats} summarizes the statistics of the datasets used in our experiments. All datasets are used without further processing except AQuA and APPS. For AQuA, we subsampled a smaller subset from the original training set while keeping the full test set unchanged. For APPS, we restricted evaluation to introductory-level problems, as harder difficulty tiers were largely intractable for models in our 3B--12B parameter range, and sampled 2,000 training and 500 evaluation examples from this subset. For every dataset, we further split the original test set into validation (20\%) and test partitions (80\%); the validation split is used exclusively for selecting the optimal centroid scale factor $\gamma$.

\noindent \textbf{Implementation Details}: All models are trained using AdamW (weight decay 0.01) with a cosine decay schedule and a 3\% linear warmup; gradients are clipped to a maximum norm of 1.0.

\begin{figure}[h]
\begin{tcolorbox}[colback=gray!5!white,colframe=gray!75!black,title=Prompt Template: Mbpp | APPs | Humaneval]
Write a python function \texttt{[Function Name]} to solve the task described below.

\textbf{Task:}
[Instruction Text]

Provide only the Python code inside a '''python''' block.
\end{tcolorbox}
\caption{The prompt template used for code generation.}
\label{fig:python_prompt}
\end{figure}

\begin{figure}[h]
\begin{tcolorbox}[colback=gray!5!white,colframe=gray!75!black,title=Prompt Template: SPIDER]
Write a SQL query to solve the task described below.

[Instruction Text]

Provide only the SQL query.
\end{tcolorbox}
\caption{The prompt template used for the SQL generation.}
\label{fig:sql_prompt}
\end{figure}

\noindent\textbf{Prompt Template}: For the math tasks, we do not use any prompt template for the training or evaluation. For the coding tasks, two different prompt templates are being used for the code generation (Figure~\ref{fig:python_prompt}) and SQL query generation (Figure~\ref{fig:sql_prompt}) separately.

\subsection{Results across Diverse Architectures}\label{subsec:full_results}
\begin{table*}[p]
\centering
\fontsize{9pt}{\baselineskip}\selectfont 
\setlength{\tabcolsep}{1mm} 
\begin{tabular}{@{}llcc@{\hspace{0.3em}}cccc@{\hspace{0.3em}}ccc@{\hspace{0.3em}}c@{}}
\toprule
 & & \multicolumn{2}{c}{\textbf{Memory (MB)}} & \multicolumn{4}{c}{\textbf{Math Reasoning Tasks}} & \multicolumn{3}{c}{\textbf{Coding Tasks}} & \\
\cmidrule(lr){3-4} \cmidrule(lr){5-8} \cmidrule(lr){9-12}
\textbf{Method} & \textbf{Rank} & \textbf{Pin} & \textbf{Swap} & \textbf{GSM8K}$^\ddagger$ & \textbf{SVAMP}$^\ddagger$ & \textbf{Multiarith}$^\dagger$ & \textbf{AQuA}$^\dagger$ & \textbf{MBPP}$^\ddagger$ & \textbf{SPIDER}$^\ddagger$ & \textbf{APPs}$^\dagger$ & \textbf{HumanEval} \\
\midrule
Base  & NA & 0.0 & 0.0 & 12.16 & 22.50 & 32.41 & 25.98 & 31.75 & 30.82 & 14.75 & 49.61 \\
\midrule
\multirow{5}{*}{LoRA} 
 & $r=1$ & 0.0 & 3.0 & 43.22 & 63.33 & 15.17 & 31.86 & 30.75 & 43.29 & 24.5 & 35.66 \\
 & $r=2$ & 0.0 & 6.1 & 43.59 & 63.75 & 24.13 & \textbf{34.80} & 31.50 & 42.45 & 23.25 & 34.88 \\
 & $r=4$ & 0.0 & 12.2 & \textbf{45.93} & 69.58 & 44.13 & \textbf{34.31} & 31.75 & 44.84 & 25.75 & 37.21 \\
 & $r=8$ & 0.0 & 24.3 & 47.15 & 73.33 & 51.72 & 34.80 & 32.00 & 44.36 & 29.25 & 35.66 \\
 & $r=16$ & 0.0 & 48.6 & 47.61 & 73.75 & 4.13 & 34.31 & 33.25 & 47.72 & 29.0 & 34.88 \\
\midrule
\multirow{3}{*}{\begin{tabular}{@{}l@{}}LoRA$^\ast$ \\ (Merged \\Dataset)\end{tabular}}
 & $r=1$ & 0.0 & 3.0 & 42.10 & 69.17 & -- & -- & 33.25 & 43.76 & -- & -- \\
 & $r=2$ & 0.0 & 6.1 & 44.25 & 68.75 & -- & -- & 33.25 & 43.40 & -- & -- \\
 & $r=4$ & 0.0 & 12.2 & 44.81 & 73.33 & -- & -- & 32.25 & 44.12 & -- & -- \\
\midrule
\multirow{3}{*}{VeRA} 
 & $r=1k$ & 111.2 & 0.8 & 27.13 & 70.00 & 66.89 & 31.37 & 4.75 & 40.33 & \textbf{38.5} & 15.50 \\
 & $r=2k$ & 222.3 & 1.6 & 28.06 & 70.00 & 72.41 & 34.80 & 4.50 & 41.01 & \textbf{41.75} & 13.95 \\
 & $r=4k$ & 444.6 & 3.2 & 28.06 & 72.50 & 70.34 & 33.82 & 4.75 & 40.60 & \textbf{44.75} & 13.95 \\
\midrule
\multirow{5}{*}{\begin{tabular}{@{}l@{}}CTS \\ (JD-Full)\end{tabular}} 
 & $r=1$ & 1.5 & 1.5 & 3.18 & 58.33 & 2.76 & 22.55 & \textbf{34.5} & \textbf{48.08}  & 35.25 & 37.98 \\
 & $r=2$ & 3.0 & 3.1 & 6.64 & 66.67 & 4.14 & 22.55 & \textbf{33.75} & \textbf{46.4} & 35.75 & 39.53 \\
 & $r=4$ & 5.9 & 6.2 & 32.93 & 71.25 & 4.82 & 32.84 & 32.25 & 42.44 & 30.0 & 41.03 \\
 & $r=8$ & 11.9 & 12.4 & 40.79 & 72.08 & 4.13 & 29.41 & 30.0 & 38.48 & 27.75 & 38.75\\
 & $r=16$ & 23.9 & 24.8 & 43.21 & 72.91 & 4.13 & 33.82 & 31.5 & 34.29 & 28.0 & 33.33\\
\midrule
\multirow{3}{*}{\begin{tabular}{@{}l@{}}\textbf{SALT} \\ \textbf{(Ours)}\end{tabular}} 
 & $r=1$ & 48.6 & 3.0 & \textbf{45.65} {\scriptsize (+2.4)} & \textbf{74.17} {\scriptsize (+10.8)} & \textbf{87.58} {\scriptsize (+72.4)} & \textbf{31.86} {\scriptsize (+0.0)} & 32.75 {\scriptsize (+2.0)} & 46.04 {\scriptsize (+2.7)} & 27.0 {\scriptsize (+2.5)} & \textbf{41.86} {\scriptsize (+6.2)} \\
 & $r=2$ & 48.6 & 6.1 & \textbf{46.02} {\scriptsize (+2.4)} & \textbf{73.33} {\scriptsize (+9.5)} & \textbf{85.51} {\scriptsize (+61.4)} &  31.86 {\scriptsize (-2.9)} & 33.00 {\scriptsize (+1.5)} & 44.72 {\scriptsize (+2.2)} & 24.25 {\scriptsize (+1.0)} & \textbf{41.08} {\scriptsize (+6.2)} \\
 & $r=4$ & 48.6 & 12.2 & 45.27 {\scriptsize (-0.6)} & \textbf{73.75} {\scriptsize (+3.7)} & \textbf{86.21} {\scriptsize (+42.1)}&  30.88 {\scriptsize (-3.4)} & \textbf{33.25} {\scriptsize (+1.5)} & \textbf{45.44} {\scriptsize (+0.6)} & 24.75 {\scriptsize (-1.0)}  & \textbf{42.63} {\scriptsize (+5.4)} \\
\bottomrule
\end{tabular}

 \raggedright
 $^\ddagger$ Datasets utilized for Phase 1 centroid training and Phase 2 residual tuning. $^\dagger$ Datasets for Phase 2 residual tuning only. 
\caption{Task performance and memory footprint across different ranks and baselines on Llama-3.2-3B.}
\label{tab:llama_result_table}
\end{table*}

\begin{table*}[p]
\centering
\fontsize{9pt}{\baselineskip}\selectfont 
\setlength{\tabcolsep}{1mm} 
\begin{tabular}{@{}llcc@{\hspace{0.3em}}cccc@{\hspace{0.3em}}ccc@{\hspace{0.3em}}c@{}}
\toprule
 & & \multicolumn{2}{c}{\textbf{Memory (MB)}} & \multicolumn{4}{c}{\textbf{Math \& Reasoning Tasks}} & \multicolumn{4}{c}{\textbf{Coding Tasks}}  \\
\cmidrule(lr){3-4} \cmidrule(lr){5-8} \cmidrule(lr){9-12} \cmidrule(lr){12-12}
\textbf{Method} & \textbf{Rank} & \textbf{Pin} & \textbf{Swap} & \textbf{GSM8K}$^\ddagger$ & \textbf{SVAMP}$^\ddagger$ & \textbf{MultiArith}$^\dagger$ & \textbf{AQuA}$^\dagger$ & \textbf{MBPP}$^\ddagger$ & \textbf{SPIDER}$^\ddagger$ & \textbf{APPs}$^\dagger$ & \textbf{HumanEval} \\
\midrule
Base & NA & -- & -- & 1.31 & 0.8  & 0.7 & 0.0 & 6.25 & 0.0 & 9.0 & 1.83 \\
\midrule
\multirow{5}{*}{LoRA} 
 & $r=1$  & 0.00 & 2.21 & 13.28 & 36.25 & 4.13 & \textbf{17.64} & 12.0 & 2.76 & 12.5 & \textbf{4.27}\\
 & $r=2$  & 0.00 & 4.42 & 13.84 & 40.0 & 7.59 & \textbf{16.67} & 11.75 &0.0 & 12.25 & \textbf{7.93} \\
 & $r=4$  & 0.00 & 8.85 & 16.56 & 41.67 & 6.21 & 18.62 & 11.75 & 3.72 & 13.25 & \textbf{2.44}\\
 & $r=8$  & 0.00 & 17.70 & 17.11 & 40.83  & 8.27 & 21.57 & 12.5 & 4.08 & 11.75 & 8.54\\
 & $r=16$ & 0.00 & 35.39 & 18.33 & 43.33 & 6.89 & 17.16 & 12.25 & 1.70 & 12.0 & 2.44\\
\midrule
\multirow{3}{*}{\begin{tabular}{@{}l@{}}LoRA$^\ast$ \\ (Merged \\Dataset)\end{tabular}}
 & $r=1$ & 0.0 & 3.0 & 12.62 & 23.75 & -- & -- & \textbf{13.0} & 0.0 & -- & -- \\
 & $r=2$ & 0.0 & 6.1 & 14.03 & 31.67 & -- & -- & 11.25 & 0.12 & -- & -- \\
 & $r=4$ & 0.0 & 12.2 & 15.34 & 31.25 & -- & -- & \textbf{15.25} & 0.0 & -- & -- \\
\midrule
\multirow{3}{*}{VeRA} 
 & $r=1024$ & 62.92 & 0.29 & 2.15 & 39.58 & 9.65 & 17.64 & 5.80 & 17.79 & 14.50 & 0.61 \\
 & $r=2048$ & 125.83 & 0.59 & 2.71 & 40.83 & 9.65 & 15.68 & 5.60 & 19.05 & \textbf{15.50} & 0.00 \\
 & $r=4096$ & 251.66 & 1.18 & 2.99 & 44.17 & 10.34 & 15.68 & 6.40 & 17.99 & \textbf{16.75} & 0.00 \\
\midrule
\multirow{5}{*}{\begin{tabular}{@{}l@{}}CTS \\ (JD-Full)\end{tabular}} 
 & $r=1$  & 0.74 & 1.48 & 1.78 & 2.92  & 4.82 & 0.0 & 11.25 & 0.0 & 9.25 &  0.0 \\
 & $r=2$  & 1.48 & 2.95 & 2.15 & 30.83 & 4.13 & 0.49 & 11.5 & 0.0  & 8.25 & 0.0\\
 & $r=4$  & 2.95 & 5.90 & 4.58 & 35.0 & 6.90 & 10.78 & 9.5 & 0.0  & 12.5 & 0.0 \\
 & $r=8$  & 5.90 & 11.80 & 9.45 & 39.58 & 6.90 & 14.22 & 11.25 & 0.0 & 11.75 & 0.0 \\
 & $r=16$ & 11.80 & 23.60 & 16.46 & 40.0 & 4.14  & 15.20 & 11.25 & 0.24 & 11.75 & 0.0 \\
\midrule
\multirow{3}{*}{\begin{tabular}{@{}l@{}}\textbf{\name{}} \\ \textbf{(Ours)}\end{tabular}} 
 & $r=1$  & 35.39 & 2.21 & \textbf{17.68} {\scriptsize (+4.4)}  & \textbf{44.58} {\scriptsize (+8.3)} & \textbf{57.24} {\scriptsize (+53.1)} & 15.69 {\scriptsize (-1.9)}& 12.25 {\scriptsize (+0.2)} & \textbf{42.20} {\scriptsize (+29.7)} & \textbf{13.25} {\scriptsize (+0.8)} & 1.83 {\scriptsize (-2.4)} \\
 & $r=2$  & 35.39 & 4.42 & \textbf{16.93}  {\scriptsize (+3.1)} & \textbf{42.5} {\scriptsize (+2.5)} & \textbf{51.72} {\scriptsize (+44.1)} & 13.73 {\scriptsize (-2.9)} & \textbf{12.0} {\scriptsize (+0.2)} & \textbf{41.72} {\scriptsize (+41.7)} & 14.25 {\scriptsize (+2.0)} & 1.22 {\scriptsize (-6.7)}  \\
 & $r=4$  & 35.39 & 8.84 & \textbf{16.65} {\scriptsize (+0.1)} & \textbf{43.33} {\scriptsize (+1.7)} & \textbf{49.66} {\scriptsize (+43.5)} & \textbf{19.60}  {\scriptsize (+1.0)} & 12.75 {\scriptsize (+1.0)}& \textbf{40.29} {\scriptsize (+36.6)} & 14.50 {\scriptsize (+1.3)} & 1.22  {\scriptsize (-1.2)}  \\
\bottomrule
\end{tabular}

$^\ddagger$ Datasets utilized for Phase 1 centroid training and Phase 2 residual tuning. $^\dagger$ Datasets for Phase 2 residual tuning only. 
\caption{Peak performance and memory footprint for Pythia-12B }
\label{tbl:pythia_table}
\end{table*}


\begin{table}[ht]
\centering
\fontsize{9pt}{\baselineskip}\selectfont 
\setlength{\tabcolsep}{1.5mm} 
\begin{tabular}{@{}lrrrr@{}}
\toprule
 & \multicolumn{2}{c}{\textbf{PCIe}} & \multicolumn{2}{c}{\textbf{VRAM}} \\
\cmidrule(lr){2-3} \cmidrule(l){4-5}
\textbf{Adapters} & \textbf{vLLM} & \textbf{vLLM+SALT} & \textbf{vLLM} & \textbf{vLLM+SALT} \\
\midrule
32  & 73.82 & 79.98\,\scriptsize{(+8.3\%)}  & 73.73 & 76.97\,\scriptsize{(+4.4\%)} \\
64  & 62.03 & 70.79\,\scriptsize{(+14.1\%)} & 62.23 & 73.14\,\scriptsize{(+17.5\%)} \\
128 & 51.65 & 65.91\,\scriptsize{(+27.6\%)} & 50.24 & 62.10\,\scriptsize{(+23.6\%)} \\
256 & 38.70 & 58.39\,\scriptsize{(+50.9\%)} & 28.60 & 36.68\,\scriptsize{(+28.3\%)} \\
\bottomrule
\end{tabular}
\caption{Throughput for Llama-3.2-3B (req/s)}
\label{tab:throughput_results_llama}
\end{table}

\begin{table}[ht]
\centering
\fontsize{9pt}{\baselineskip}\selectfont 
\setlength{\tabcolsep}{1.5mm} 
\begin{tabular}{@{}lrrrr@{}}
\toprule
 & \multicolumn{2}{c}{\textbf{PCIe}} & \multicolumn{2}{c}{\textbf{VRAM}} \\
\cmidrule(lr){2-3} \cmidrule(l){4-5}
\textbf{Adapters} & \textbf{vLLM} & \textbf{vLLM+SALT} & \textbf{vLLM} & \textbf{vLLM+SALT} \\
\midrule
16  & 64.20 & 65.36\,\scriptsize{(+1.8\%)}  & 11.27 & 11.53\,\scriptsize{(+2.3\%)} \\
32  & 62.67 & 64.74\,\scriptsize{(+3.3\%)}  & 10.47 & 11.18\,\scriptsize{(+6.8\%)} \\
64  & 56.51 & 55.78\,\scriptsize{(-1.3\%)}  & 8.41  & 9.79\,\scriptsize{(+16.4\%)}  \\
128 & 47.40 & 53.89\,\scriptsize{(+13.7\%)} & 5.11  & 7.27\,\scriptsize{(+42.3\%)}  \\
\bottomrule
\end{tabular}
\caption{Throughput for Pythia-12B (req/s).}
\label{tab:throughput_results_pythia}
\end{table}
\begin{figure}[h]
  \centering
  \includegraphics[width=1.0\columnwidth]{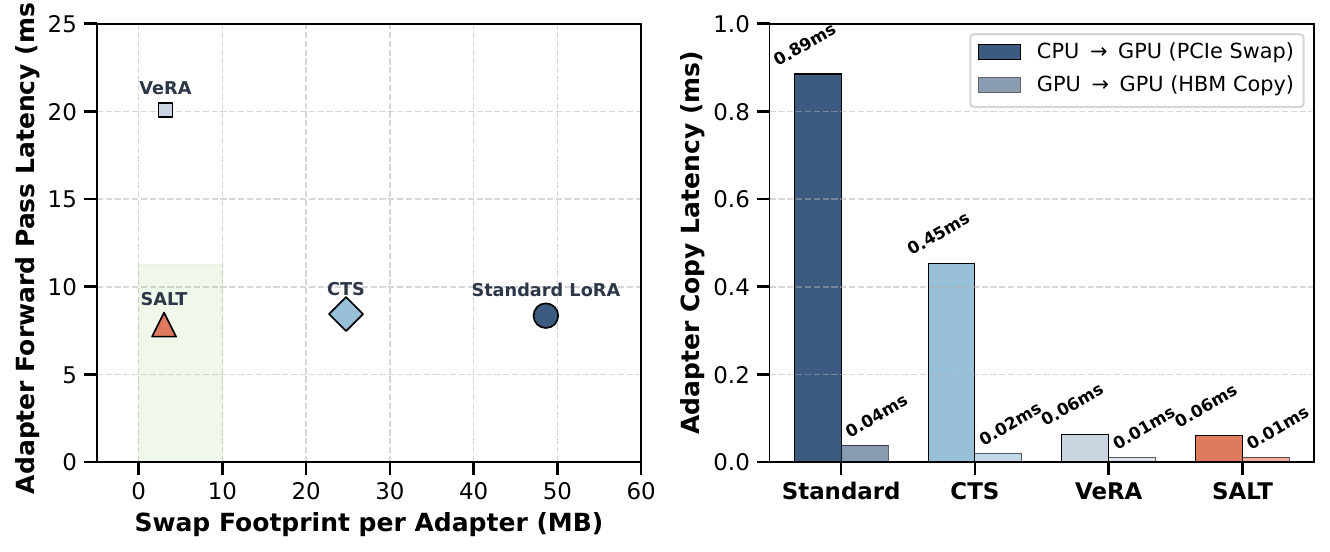} 
  \caption{Llama-3.2-3B: (Left) Swap footprint versus adapter forward pass latency. (Right) Latency breakdown for PCIe swap and HBM copy operations across different baselines.}
  \label{fig:micro_bench_llama}
\end{figure}

\begin{figure}[h]
  \centering
  \includegraphics[width=1.0\columnwidth]{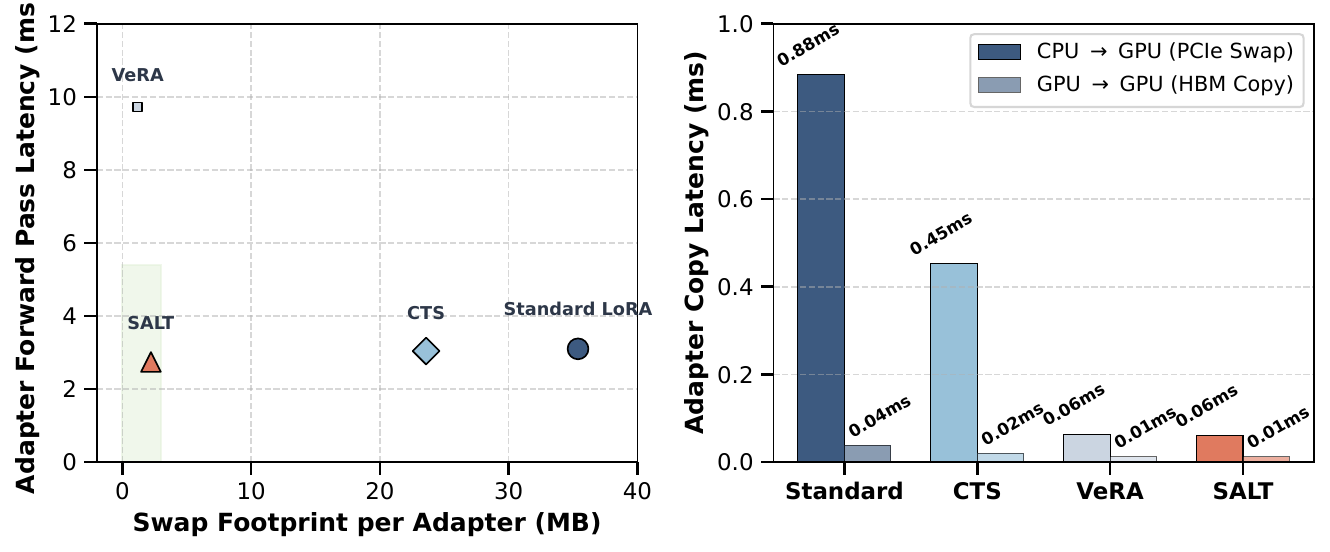} 
  \caption{Pythia-12B: (Left) Swap footprint versus adapter forward pass latency. (Right) Latency breakdown for PCIe swap and HBM copy operations across different baselines.}
  \label{fig:micro_bench_pythia}
\end{figure}

\begin{figure*}[h]
\begin{lstlisting}[caption={Representative failure case: the naively concatenated centroid answers correctly on its own, but a Phase 2 residual trained on top of it hallucinate an unnecessary table join.}, label={lst:sql_failure_cases}]
/* =================================================================
   QUESTION: Return the document id, template id, and description for the document with the name Robbin CV.
   ================================================================= */
-- [1] GOLD QUERY (Reference Answer)
SELECT document_id, template_id, Document_Description FROM Documents WHERE document_name = "Robbin CV";
-- [2] CONCATENATED CENTROID ONLY, NO RESIDUAL (Correct)
SELECT document_id, template_id, document_description FROM Documents WHERE document_name = "Robbin CV";
-- [3] CONCATENATED CENTROID + RESIDUAL (Failed)
SELECT T1.document_id, T1.template_id, (*@\textcolor{errorred}{\textbf{T2.}}@*)document_description FROM Documents AS T1 (*@\textcolor{errorred}{\textbf{JOIN Paragraphs AS T2}}@*) 
  ON T1.document_id = (*@\textcolor{errorred}{\textbf{T2.}}@*)document_id WHERE (*@\textcolor{errorred}{\textbf{T2.}}@*)document_name = "Robbin CV";
/* -----------------------------------------------------------------
   ERROR: Adding the residual hallucinates an unnecessary JOIN with the "Paragraphs" table and incorrectly aliases schema columns, overcomplicating a simple single-table lookup.
   ----------------------------------------------------------------- */
\end{lstlisting}
\end{figure*}

To verify that the performance gains and efficiency of \name\ are not uniquely tied to a specific model family, we extend our evaluation to two additional models with distinct architectural designs and parameter scales: Llama-3.2-3B and Pythia-12B. To accommodate their diverse architectures, we adjusted the LoRA target modules accordingly. Unlike Mistral-7B-v0.3, where we only target the \texttt{q\_proj} and \texttt{v\_proj} modules, we target all linear layers for Llama-3.2-3B and Pythia-12B.

Following the two-phase training recipe described in Section~\ref{sec:methods}, we independently trained mathematical and coding centroids for both models, followed by task-specific residual fine-tuning in Phase~2. Detailed downstream accuracies for \name\ on these architectures are presented in Tables~\ref{tab:llama_result_table} and~\ref{tbl:pythia_table}. \name\ demonstrates competitive, and in several cases superior, performance compared to standard LoRA at higher ranks, despite relying on a shared subspace and an ultra-low-rank task residual. While standard LoRA occasionally achieves higher accuracy on specific tasks at higher ranks (e.g., AQuA on Llama, GSM8K on Pythia), this is expected as standard LoRA dedicates its entire parameter budget to a single task, whereas \name\ routes task representations through a unified centroid to enable efficient multi-tenant serving. 

Similar to our primary evaluation methodology, we conducted microbenchmarks and vLLM throughput tests to profile memory overhead and inference latency for \name\ against all baselines on these models (Figures~\ref{fig:micro_bench_llama} and~\ref{fig:micro_bench_pythia}). \name\ achieves either the lowest or second-lowest adapter memory footprint and the lowest inference latency, while matching baseline downstream accuracy. As shown in Tables~\ref{tab:throughput_results_llama} and~\ref{tab:throughput_results_pythia}, plugging \name\ directly into vLLM yields throughput improvements of up to 50.9\% under PCIe bandwidth pressure for Llama-3.2-3B and up to 42.3\% under HBM capacity contention for Pythia-12B.

\subsection{Multi-Domain Task Interference}\label{subsec:mix_domain}

 We compare the performance of task residuals fine-tuned over a domain-specific (Math) centroid and a multi-domain (Math + Code) centroid while fixing the centroid rank to 16. As shown in Table~\ref{tab:centroid_negative_transfer}, integrating more data distributions into a single Phase 1 centroid may lead to degrading performance due to negative transfer. 

 \begin{table}[H]
\centering
\fontsize{9pt}{\baselineskip}\selectfont
\setlength{\tabcolsep}{5pt}
\begin{tabular}{@{}lcc@{}}
\toprule
 & \textbf{GSM8K} & \textbf{SVAMP} \\
\midrule
2-Tasks & 46.65 & 74.17 \\
4-Tasks & 44.03 & 72.33 \\
6-Tasks & 43.92 & 70.0 \\
\textbf{$\Delta$} & $-$2.73 & $-$4.17 \\
\bottomrule
\end{tabular}
\caption{Impact of multi-domain centroid training on downstream task performance. Expanding the shared centroid (Math) to include out-of-domain tasks (Code and Language) degrade the performance of downstream tasks.}
\label{tab:centroid_negative_transfer}
\end{table}

\subsection{Training Resource Complexity}\label{subsec:training_src}

\begin{table}[ht]
\centering
{ 
\fontsize{9pt}{\baselineskip}\selectfont
\setlength{\tabcolsep}{2.5mm}
\begin{tabular}{@{}lrrrrr@{}}
\toprule
\# Adapters ($M$)  & 10 & 20 & 30 & 40 & 50 \\
\midrule
Peak HBM (GB)      & 31.14 & 34.76 & 38.38 & 42.01 & 45.63 \\
Step Latency (s)   & 2.90  & 6.07  & 9.86  & 13.49 & 18.24 \\
\bottomrule
\end{tabular}
} 
\caption{Phase 1 joint alignment training scalability on Llama-3.2-3B on an NVIDIA H100.}
\label{tab:training_scalability}
\end{table}

In Phase 1, \name\ jointly trains a shared centroid with $M$ standard ($r=16$) task adapters in the same domain. The activation memory stays bounded because adapters are processed sequentially, and each compute graph is freed immediately after its backward pass. However, the weights and optimizer states for all $M$ adapters remain resident throughout training, scales as $\mathcal{O}(M)$, growing by $\sim$360~MB per adapter (Table~\ref{tab:training_scalability}) for Llama-3.2-3B adapter targeting all linear layers. Per-step latency scales similarly. Even at $M=50$, a single training step completes in $\sim$18.2s and fits comfortably within the 80~GB memory capacity of a single H100 GPU.

\subsection{Effect of Subspace-Aligned Training}\label{subsec:full_subspace_analysis}
\begin{table}[H]
\centering
\fontsize{9pt}{\baselineskip}\selectfont
\setlength{\tabcolsep}{1mm} 
\begin{tabular}{@{}llcccc@{}}
\toprule
 & & \multicolumn{4}{c}{\textbf{Accuracy \%}} \\
\cmidrule(l){3-6}
\textbf{Rank} & \textbf{Centroid Method} & \textbf{GSM8K} & \textbf{SVAMP} & \textbf{MBPP} & \textbf{SPIDER} \\
\midrule
$r=1$ & Concatenated & 52.10 & 64.17 & 22.0 & 7.07 \\
 & Subspace-aligned & \textbf{56.88} & \textbf{69.17} & \textbf{27.25} & \textbf{56.69} \\
\midrule
$r=2$ & Concatenated & 52.94 & 63.75 & 22.5 & 12.47 \\
 & Subspace-aligned & \textbf{57.81} & \textbf{69.17} & \textbf{27.8} & \textbf{56.35} \\
\midrule
$r=4$ & Concatenated & 52.67 & 65.00 & 22.75 & 12.11 \\
 & Subspace-aligned & \textbf{53.23} & \textbf{68.75} & \textbf{27.25} & \textbf{56.47} \\
\bottomrule
\end{tabular}
\caption{Effect of subspace-aligned centroid training on downstream task performance for Mistral-7B-v0.3.}
\label{tab:ablation_centroid_combined_mistral}
\end{table}

Table~\ref{tab:ablation_centroid_combined_mistral} shows the full results underlying the paired $t$-test in Section~\ref{subsec:sub_t_test}. Spider shows a distinct failure pattern. The LoRA (Merged Dataset) baseline in Table~\ref{tbl:mistral_table}, where the centroid trained on concatenated data, without any Phase 2 residual, already achieves $\sim$50\% accuracy, yet adding a Phase 2 residual on top of this same centroid degrades performance to as low as 7.07\% at $r=1$. Accuracy also remains unstable across rank for this configuration (7.07--12.47\%), whereas the residual trained on the frozen  subspace-aligned centroid consistently achieve high performance (56.35--56.69\%) regardless of residual rank. Listing~\ref{lst:sql_failure_cases} shows an example of this failure. The concatenated centroid alone answers correctly, but training a residual on top of it causes the residual to hallucinate a query (i.e. JOIN Paragraphs AS T2) rather than refine the centroid's correct output. This suggests alignment is a precondition for safe residual composition: on an unaligned centroid, the residual's limited capacity produces arbitrary structural errors rather than refinements. 

\subsection{Fine-grained Sweep of $\gamma$}
As introduced in Section~\ref{subsec:multi-serving}, \name\ discretizes the inference-time scaling coefficient $\gamma$ into predefined bins to group requests that share the same centroid and enable efficient centroid fusion. To empirically validate the practicality of this design, we performed a fine-grained sweep of $\gamma \in [0.2, 1.5]$ with a step size of 0.1, measuring downstream performance across seven datasets on Mistral-7B-v0.3.
\begin{figure}[h]
  \centering
  \includegraphics[width=1.0\columnwidth]{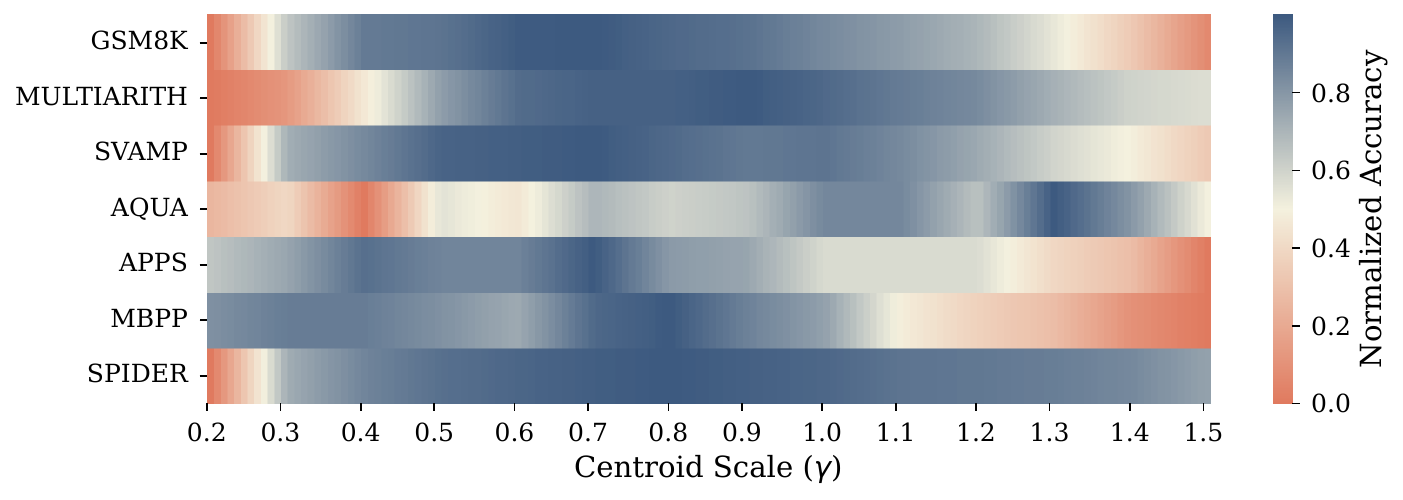} 
  \caption{Normalized downstream performance across a fine-grained sweep of the centroid scale factor $\gamma$.}
  \label{fig:gamma_fine_grain}
\end{figure}

Figure~\ref{fig:gamma_fine_grain} shows that three of the four mathematical reasoning datasets achieve optimal performance within the $0.6$--$0.9$ range before degrading sharply. AQuA remains an outlier, requiring a higher $\gamma$ (near 1.3) to fully recover accuracy. Conversely, all three coding datasets reach peak accuracy when $\gamma \in [0.7, 0.9]$. Crucially, within this $0.6$--$0.9$ optimal band, the average performance drop from the absolute peak is merely 3.23\% for math tasks and 1.53\% for coding tasks. This stability empirically confirms that adopting a small set of shared $\gamma$ bins (e.g., $\gamma \in \{0.6, 0.7 0.8, 0.9\}$) effectively optimizes the downstream task performance while improving the serving throughput.

\end{document}